# Camera Style Adaptation for Person Re-identification

Zhun Zhong[†§], Liang Zheng[§], Zhedong Zheng[§], Shaozi Li[†*], Yi Yang[§]

[†]Cognitive Science Department, Xiamen University, China
[§] Centre for Artificial Intelligence, University of Technology Sydney, Australia
{zhunzhong007,liangzheng06,zdzheng12,yee.i.yang}@gmail.com szlig@xmu.edu.cn

## Abstract

*Being a cross-camera retrieval task, person re-identification suffers from image style variations caused by different cameras. The art implicitly addresses this problem by learning a camera-invariant descriptor subspace. In this paper, we explicitly consider this challenge by introducing camera style (CamStyle) adaptation. CamStyle can serve as a data augmentation approach that smooths the camera style disparities. Specifically, with CycleGAN, labeled training images can be style-transferred to each camera, and, along with the original training samples, form the augmented training set. This method, while increasing data diversity against over-fitting, also incurs a considerable level of noise. In the effort to alleviate the impact of noise, the label smooth regularization (LSR) is adopted. The vanilla version of our method (without LSR) performs reasonably well on few-camera systems in which over-fitting often occurs. With LSR, we demonstrate consistent improvement in all systems regardless of the extent of over-fitting. We also report competitive accuracy compared with the state of the art. Code is available at:*
<https://github.com/zhunzhong07/CamStyle>

## 1. Introduction

Person re-identification (re-ID) [43] is a cross-camera retrieval task. Given a query person-of-interest, it aims to retrieve the same person from a database collected from multiple cameras. In this task, a person image often undergoes intensive changes in appearance and background. Capturing images by different cameras is a primary cause of such variations (Fig. 1). Usually, cameras differ from each other regarding resolution, environment illumination, *etc*.

In addressing the challenge of camera variations, a previous body of the literature chooses an implicit strategy. That is, to learn stable feature representations that have invari-ance property under different cameras. Examples in traditional approaches include KISSME [16], XQDA [20], DNS [39], *etc*. Examples in deep representation learning methods include IDE [43], SVDNet [29], TripletNet [11], *etc*.

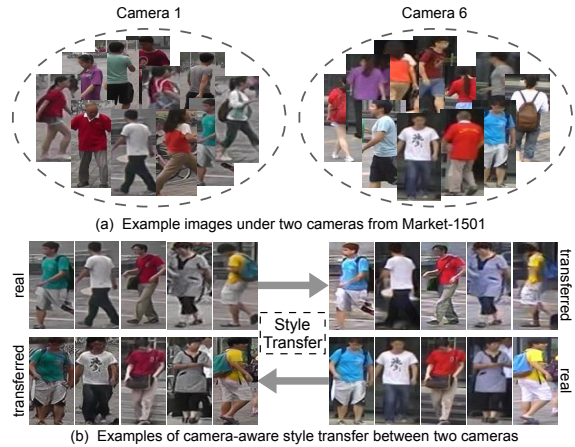

Figure 1. (a) Example images from Market-1501 [42]. (b) Examples of camera-aware style transfer between two cameras using our method. Images in the same column represent the same person.

Comparing to previous methods, this paper resorts to an explicit strategy from the view of camera style adaptation. We are mostly motivated by the need for large data volume in deep learning based person re-ID. To learn rich features which are robust to camera variations, annotating large-scale datasets is useful but prohibitively expensive. Nevertheless, if we can add more samples to the training set that are aware of the style differences between cameras, we are able to 1) address the data scarcity problem in person re-ID and 2) learn invariant features across different cameras. Preferably, this process should not cost any more human labeling, so that the budget is kept low.

Based on the above discussions, we propose a camera style (CamStyle) adaptation method to regularize CNN training for person re-ID. In its *vanilla version*, we learn image-image translation models for each camera pair with CycleGAN [51]. With the learned CycleGAN model, for a training image captured by a certain camera, we can gener-ate new training samples in the style of other cameras. In this manner, the training set is a combination of the original training images and the style-transferred images. The style-transferred images can directly borrow the label from the original training images. During training, we use the new training set for re-ID CNN training following the baseline model in [43]. The vanilla method is beneficial in reducing over-fitting and achieving camera-invariant property, but, importantly, we find that it also introduces noise to the system (Fig. 2). This problem deteriorates its benefit under full-camera systems where the relatively abundant data has a lower over-fitting risk. To mitigate this problem, in the *improved version*, we further apply label smoothing regularization (LSR) [30] on the style-transferred samples, so that their labels are softly distributed during training.

---

[*]Corresponding author.
This work was done when Zhun Zhong was a visiting student at University of Technology Sydney.

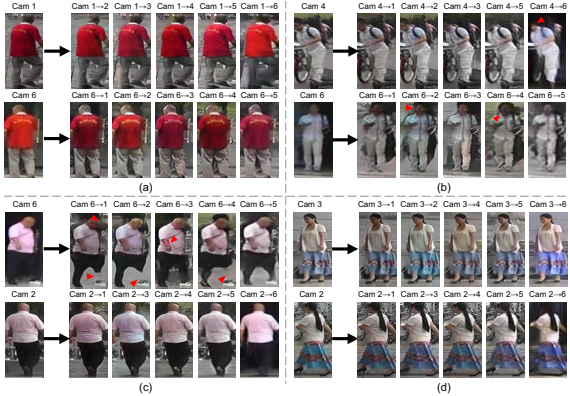

Figure 2. Examples of style-transferred samples in Market-1501 [42]. An image captured in a certain camera is translated to styles in other 5 cameras. Despite the success cases, image-image translation noise indicated by red arrows should be considered.

The proposed camera style adaptation approach, CamStyle, has three advantages. First, it can be regarded as a data augmentation scheme that not only smooths the camera style disparities, but also reduces the impact of CNN over-fitting. Second, by incorporating camera information, it helps learn pedestrian descriptors with the camera-invariant property. Finally, it is unsupervised, guaranteed by CycleGAN, indicating fair application potentials. To summarize, this paper has the following contributions:

- A vanilla camera-aware style transfer model for re-ID data augmentation. In few-camera systems, the improvement can be as large as 17.1%.

- An improved method applying LSR on the style-transferred samples during re-ID training. In full-camera systems, consistent improvement is observed.

## 2. Related Work

**Deep learning person re-identification.** Many deep learning methods [38, 34, 33, 3, 24] have been proposed in person re-ID. In [38], input image pairs are partitioned into three overlapping horizontal parts respectively, and through a siamese CNN model to learn the similarity of them using cosine distance. Later, Wu *et al.* [34] increase the depth of networks with using smaller convolution filters to obtain a robust feature. In addition, Varior *et al.* [33] merge long short-term memory (LSTM) model into a siamese network that can handle image parts sequentially so that the spatial information can be memorized to enhance the discriminative capability of the deep features.

Another effective strategy is the classification model, which makes full use of the re-ID labels [43, 35, 29, 18, 36, 44, 41]. Zheng *et al.* [43] propose the ID-discriminative embedding (IDE) to train the re-ID model as image classification which is fine-tuned from the ImageNet [17] pre-trained models. Wu *et al.* [35] propose a Feature Fusion Net (FFN) by incorporating hand-crafted features into CNN features. Recently, Sun *et al.* [29] iteratively optimize the fully connected (FC) feature with Singular Vector Decomposition and produce orthogonal weights.

When a CNN model is excessively complex compared to the number of training samples, over-fitting might happen. To address this problem, several data augmentation and regularization methods have been proposed. In [23], Niall *et al.* improve the generalization of network by utilizing background and linear transformations to generate various samples. Recently, Zhong *et al.* [49] randomly erase a rectangle region in input image with random values which prevents the model from over-fitting and makes the model robust to occlusion. Zhu *et al.* [50] randomly select Pseudo-Positive samples from an independent dataset as addition training samples for training re-ID CNN to reduce the risk of over-fitting. More related to this work, Zheng *et al.* [47] use DCGAN [25] to generate unlabeled samples, and assign them with a uniform label distribution to regularize the network. In contrast to [47], the style-transferred samples in this work are produced from real data with relatively reliable labels.

**Generative Adversarial Networks.** Generative Adversarial Networks (GANs) [9] have achieved impressive success in recent years, especially in image generation [25]. Recently, GANs have also been applied to image-to-image translation [13, 51, 22], style transfer [8, 14, 6] and cross domain image generation [2, 31, 5]. Isola *et al.* [13] apply a conditional GANs to learn a mapping from input to output images for image-to-image translation application. The main drawback of [13] is that it requires pairs of corresponding images as training data. To overcome this problem, Liu and Tuzel [22] propose a coupled generative adversarial network (CoGAN) by employing weight-sharing networks to learn a joint distribution across domains. More recently, CycleGAN [51] introduces cycle consistency based on "pix2pix" framework in [13] to learn the image trans-lation between two different domains without paired samples. Style transfer and cross domain image generation can also be regarded as image-to-image translation, in which the style (or domain) of input image is transferred to another while remaining the original image content. In [8], a style transfer method is introduced by separating and recombining the content and style of images. Bousmalis *et al.* [2] introduce an unsupervised GAN framework that transfer images from source domain to an analog image in target domain. Similarity, in [31], the Domain Transfer Network (DTN) is proposed by incorporating multiclass GAN loss to generate images of unseen domain, while reserving original identity. Unlike previous methods which mainly consider the quality of the generated samples, this work aims at using the style-transferred samples to improve the performance of re-ID.

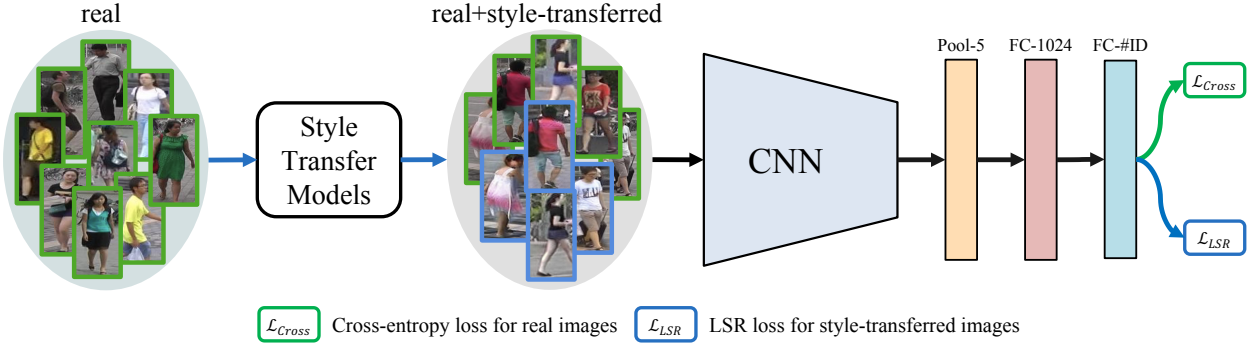

Figure 3. The pipeline of our method. The camera-aware style transfer models are learned from the real training data between different cameras. For each real image, we can utilize the trained transfer model to generate images which fit the style of target cameras. Subsequently, real images (green boxes) and style-transferred images (blue boxes) are combined to train the re-ID CNN. The cross-entropy loss and the label smooth regularization (LSR) loss are applied to real images and style-transferred images, respectively.

## 3. The Proposed Method

In this section, we first briefly look back at the CycleGAN [51] in Section 3.1. We then describe the camera-aware data generation process using CycleGAN in Section 3.2. The baseline and the training strategy with LSR are described in Section 3.3 and Section 3.4, respectively. The overall framework is shown in Fig. 3.

### 3.1. CycleGAN Review

Given two datasets $\{x_i\}_{i=1}^{\mathcal{M}}$ and $\{y_j\}_{j=1}^{\mathcal{N}}$, collected from two different domains $A$ and $B$, where $x_i \in A$ and $y_j \in B$, The goal of CycleGAN is to learn a mapping function $G : A \to B$ such that the distribution of images from $G(A)$ is indistinguishable from the distribution $B$ using an adversarial loss. CycleGAN contains two mapping functions $G : A \to B$ and $F : B \to A$. Two adversarial discriminators $D_A$ and $D_B$ are proposed to distinguish whether images are translated from another domain. CycleGAN applies the GAN framework to jointly train the generative and discriminative models. The overall CycleGAN loss function is expressed as:

$$V(G, F, D_A, D_B) = V_{GAN}(D_B, G, A, B) \\ + V_{GAN}(D_A, F, B, A) \quad (1) \\ + \lambda V_{cyc}(G, F),$$

where $V_{GAN}(D_B, G, A, B)$ and $V_{GAN}(D_A, F, B, A)$ are the loss functions for the mapping functions $G$ and $F$ and for the discriminators $D_B$ and $D_A$. $V_{cyc}(G, F)$ is the *cycle consistency loss* that forces $F(G(x)) \approx x$ and $G(F(y)) \approx y$, in which each image can be reconstructed after a cycle mapping. $\lambda$ penalizes the importance between $V_{GAN}$ and $V_{cyc}$. More details about CycleGAN can be accessed in [51].

### 3.2. Camera-aware Image-Image Translation

In this work, we employ CycleGAN to generate new training samples: the styles between different cameras are considered as different domains. Given a re-ID dataset containing images collected from $L$ different camera views, our method is to learn image-image translation models for each camera pair with CycleGAN. To encourage the style-transfer to preserve the color consistency between the input and output, we add the *identity mapping loss* [51] in the CycleGAN loss function (Eq. 1) to enforce the generator to approximate an identity mapping when using real images of the target domain as input. The *identity mapping loss* can be expressed as:

$$V_{identity}(G, F) = E_{x \sim p_x}[\|F(x) - x\|_1] \\ + E_{y \sim p_y}[\|G(y) - y\|_1], \quad (2)$$

Specifically, for training images, we use CycleGAN to train camera-aware style transfer models for each pair of cameras. Following the training strategy in [51], all images are resized to $256 \times 256$. We use the same architecture for our camera-aware style transfer networks as CycleGAN. The generator contains 9 residual blocks and four convolutions, while the discriminator is $70 \times 70$ PatchGANs [13].

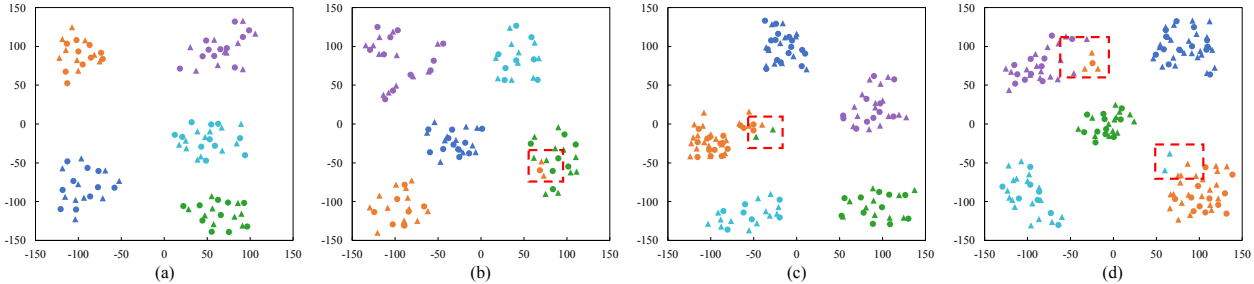

Figure 4. Barnes-Hut t-SNE [32] visualization on Market-1501. We randomly select real training images of 700 identities to train the re-ID model and visualize the real samples (R, dots) and their fake (style-transferred) samples (F, triangles) of a rest 20 identities. In each figure, different colors represent different identities. We observe 1) fake samples generally overlay with the real samples, laying the foundation of their data augmentation mechanism; 2) noisy fake data exist now and then (in red boxes), which needs regularization techniques such as LSR. Best viewed in color.

With the learned CycleGAN models, for a training image collected from a certain camera, we generate $L-1$ new training samples whose styles are similar to the corresponding cameras (examples are shown in Fig. 2). In this work, we call the generated image as **style-transferred** image or **fake** image. In this manner, the training set is augmented to a combination of the original images and the style-transferred images. Since each style-transferred image preserves the content of its original image, the new sample is considered to be of the same identity as the original image. This allows us to leverage the style-transferred images as well as their associated labels to train re-ID CNN in together with the original training samples.

**Discussions.** As shown in Fig. 4, the working mechanism of the proposed data augmentation method mainly consists in: 1) the similar data distribution between the real and fake (style-transferred) images, and 2) the ID labels of the fake images are preserved. In the first aspect, the fake images fill up the gaps between real data points and marginally expand the class borders in the feature space. This guarantees that the augmented dataset generally supports a better characterization of the class distributions during embedding learning. The second aspect, on the other hand, supports the usage of supervised learning [43], a different mechanism from [47] which leverages unlabeled GAN images for regularization.

### 3.3. Baseline Deep Re-ID Model

Given that both the real and fake (style-transferred) images have ID labels, we use the ID-discriminative embedding (IDE) [43] to train the re-ID CNN model. Using the Softmax loss, IDE regards re-ID training as an image classification task. We use ResNet-50 [10] as backbone and follow the training strategy in [43] for fine-tuning on the ImageNet [4] pre-trained model. Different from the IDE proposed in [43], we discard the last 1000-dimensional classification layer and add two fully connected (FC) layers. The output of the first FC layer has 1024 dimensions named as "FC-1024", followed by batch normalization [12], ReLU and Dropout [27]. The addition "FC-1024" follows the practice in [29] which yields improved accuracy. The output of the second FC layer, is $C$-dimensional, where $C$ is the number of classes in the training set. In our implementation, all input images are resized to $256 \times 128$. The network is illustrated in Fig. 3.

### 3.4. Training with CamStyle

Given a new training set composed of real and fake (style-transferred) images (with their ID labels), this section discusses the training strategies using the CamStyle. When we view the real and fake images equally, *i.e.,* assigning a "one-hot" label distribution to them, we obtain a *vanilla version* of our method. On the other hand, when considering the noise introduced by the fake samples, we introduce the *full version* which includes the label smooth regularization (LSR) [30].

**Vanilla version.** In the vanilla version, each sample in the new training set belongs to a single identity. During training, in each mini-batch, we randomly select $M$ real images and $N$ fake images. The loss function can be written as,

$$\mathcal{L} = \frac{1}{M}\sum_{i=1}^{M}\mathcal{L}_R^i + \frac{1}{N}\sum_{j=1}^{N}\mathcal{L}_F^j, \quad (3)$$

where $\mathcal{L}_R$ and $\mathcal{L}_F$ are the cross-entropy loss for real images and fake images, respectively. The cross-entropy loss function can be formulated as,

$$\mathcal{L}_{Cross} = -\sum_{c=1}^{C}\log(p(c))q(c), \quad (4)$$

where $C$ is the number of classes, and $p(c)$ is the predicted probability of the input belonging to label $c$. $p(c)$ is normalized by the softmax layer, so $\sum_{c=1}^{C}p(c) = 1$. $q(c)$ is the ground-truth distribution. Since each person in the training set belongs to a single identity $y$, $q(c)$ can be defined as,

$$q(c) = \begin{cases} 1 & c = y \\ 0 & c \neq y. \end{cases} \quad (5)$$

So minimizing the cross entropy is equivalent to maximizing the probability of the ground-truth label. For a given person with identity $y$, the cross-entropy loss in Eq. 4 can be rewritten as,

$$\mathcal{L}_{Cross} = -\log p(y). \quad (6)$$

Because the similarity in overall data distribution between the real and fake data, the vanilla version is able to improve the baseline IDE accuracy under a system with a few cameras, as to be shown in Section 4.

**Full version.** The style-transferred images have a positive data augmentation effect, but also introduce noise to the system. Therefore, while the vanilla version has merit in reducing over-fitting under a few-camera system in which, due to the lack of data, over-fitting tends to occur, its effectiveness is compromised under more cameras. The reason is that when data from more cameras is available, the over-fitting problem is less critical, and the problem of transfer noise begins to appear.

The transfer noise arises from two causes. 1) CycleGAN does not perfectly model the transfer process, so errors occur during image generation. 2) Due to occlusion and detection errors, there exists noisy samples in the real data, transferring these noisy samples to fake data may produce even more noisy samples. In Fig. 4, we visualize some examples of the deep feature of real and fake data on a 2-D space. Most of the generated samples are distributed around the original images. When transfer errors happen (see Fig. 4(c) and Fig. 4(d)), the fake sample will be a noisy sample and be far away from the true distribution. When a real image is a noise sample (see Fig. 4(b) and Fig. 4(d)), it is far away from the images with the same labels, so its generated samples will also be noisy. This problem reduces the benefit of generated samples under full-camera systems where the relatively abundant data has a lower over-fitting risk.

To alleviate this problem, we apply the label smoothing regularization (LSR) [30] on the style-transferred images to softly distribute their labels. That is, we assign less confidence on the ground-truth label and assign small weights to the other classes. The re-assignment of the label distribution of each style-transferred image is written as,

$$q_{LSR}(c) = \begin{cases} 1 - \epsilon + \frac{\epsilon}{C} & c = y \\ \frac{\epsilon}{C} & c \neq y, \end{cases} \quad (7)$$

where $\epsilon \in [0, 1]$. When $\epsilon = 0$, Eq. 7 can be reduced to

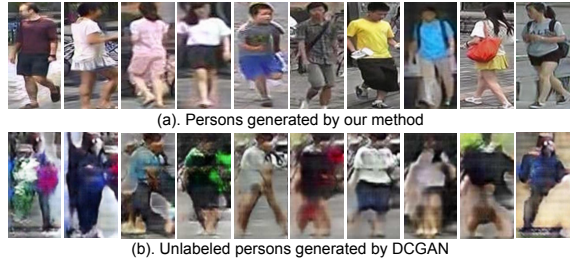

(a). Persons generated by our method

(b). Unlabeled persons generated by DCGAN

Figure 5. Examples generated by our method and DCGAN in [47].

Eq. 5. Then, the cross-entropy loss in Eq. 4 is re-defined as,

$$\mathcal{L}_{LSR} = -(1-\epsilon)\log p(y) - \frac{\epsilon}{C}\sum_{c=1}^{C}\log p(c) \quad (8)$$

For real images, we do not use LSR because their labels correctly match the image content. Moreover, we experimentally show that adding LSR to the real images does not improve the re-ID performance under full-camera systems (see Section 4.4). So for real images, we use the one-hot label distribution. For style-transferred images, we set $\epsilon = 0.1$, the loss function $\mathcal{L}_F = \mathcal{L}_{LSR}(\epsilon = 0.1)$.

**Discussions.** Recently, Zheng *et al.* [47] propose the label smoothing regularization for outliers (LSRO) to use the unlabeled samples generated by DCGAN [25]. In [47], since the generated images do not have labels, a uniform label distribution is assigned to the generated samples, *i.e.*, $\mathcal{L}_{LSR}(\epsilon = 1)$. Comparing with LSRO [47], our system has two differences. 1) Fake images are generated according to camera styles. The usage of CycleGAN ensures that the generated images remain the main characteristics of the person (Fig. 5 provides some visual comparisons). 2) Labels in our systems are more reliable. We use LSR to address a small portion of unreliable data, while LSRO [47] is used under the scenario where no labels are available.

## 4. Experiment

### 4.1. Datasets

We evaluate our method on Market-1501 [42] and DukeMTMC-reID [47, 26], because both datasets 1) are large-scale and 2) provide camera labels for each image.

**Market-1501** [42] contains 32,668 labeled images of 1,501 identities collected from 6 camera views. Images are detected using deformable part model [7]. The dataset is split into two fixed parts: 12,936 images from 751 identities for training and 19,732 images from 750 identities for testing. There are on average 17.2 images per identity in the training set. In testing, 3,368 hand-drawn images from 750 identities are used as queries to retrieve the matching persons in the database. Single-query evaluation is used.

**DukeMTMC-reID** [47] is a newly released large-scale person re-ID dataset. It is collected from 8 cameras and comprised of 36,411 labeled images belonging to 1,404 identities. Similar to Market-1501, it consists of 16,522 training images from 702 identities, 2,228 query images from the other 702 identities and 17,661 database images. We use rank-1 accuracy and mean average precision (mAP) for evaluation on both datasets.

### 4.2. Experiment Settings

**Camera-aware style transfer model.** Following Section 3.2, given a training set captured from $L$ camera views, we train a camera-aware style transfer (CycleGAN) model for each pair of cameras. Specifically, we train $C_6^2 = 15$ and $C_8^2 = 28$ CycleGAN models for Market-1501 and DukeMTMC-reID, respectively. During training, we resize all input images to $256 \times 256$ and use the Adam optimizer [15] to train the models from scratch with $\lambda = 10$ for all the experiments. We set the batch size = 1. The learning rate is 0.0002 for the Generator and 0.0001 for the Discriminator at the first 30 epochs and is linearly reduced to zero in the remaining 20 epochs. In camera-aware style transfer step, for each training image, we generated $L - 1$ (5 for Market-1501 and 7 for DukeMTMC-reID) extra fake training images with their original identity preserved as augmented training data.

**Baseline CNN model for re-ID.** To train the baseline, we follow the training strategy in [43]. Specifically, we keep the aspect ratio of all images and resize them to $256 \times 128$. Two data augmentation methods, random cropping and random horizontal flipping are employed during training. The dropout probability $p$ is set to 0.5. We use ResNet-50 [10] as backbone, in which the second fully connected layer has 751 and 702 units for Market-1501 and DukeMTMC-reID, respectively. The learning rate starts with 0.01 for ResNet-50 base layers and 0.1 for the two new added full connected layers. We use the SGD solver to train re-ID model and set the batch size to 128. The learning rate is divided by 10 after 40 epochs, we train 50 epochs in total. In testing, we extract the output of the Pool-5 layer as image descriptor (2,048-dim) and use the Euclidean distance to compute the similarity between images.

**Training CNN with CamStyle.** We use the same setting as training the baseline model, except that we randomly select $M$ real images and $N$ fake (style-transferred) images in a training mini-batch. If not specified, we set $M : N = 3 : 1$. Note that, since the number of fake images is larger than that of real images, in each epoch, we use all the real images and randomly selected a $\frac{N}{M} \times \frac{1}{L-1}$ proportion of all fake images.

### 4.3. Parameter Analysis

An important parameter is involved with CamStyle, *i.e.*, the ratio of $\frac{M}{N}$, where $M$ and $N$ indicate the number of real and fake (style-transferred) training samples in the mini-batch. This parameter encodes the fraction of fake samples used in training. By varying this ratio, we show the experimental results in Fig. 6. It can be seen that, CamStyle with different $\frac{M}{N}$ consistently improves over the baseline. When using more fake data than real data ($M : N < 1$) in each mini-batch, our approach slightly gains about 1% improvement in rank-1 accuracy. On the contrary, when $M : N > 1$, our approach yields more than 2% improvement in rank-1 accuracy. The best performance is achieved when $M : N = 3 : 1$.

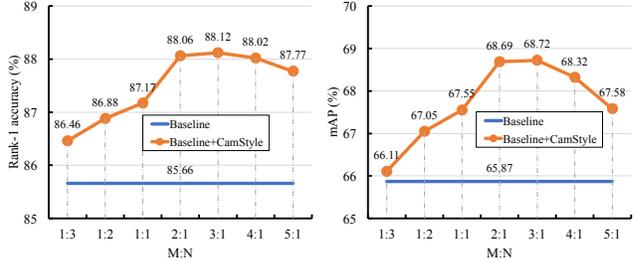

Figure 6. Evaluation with different ratio of real data and fake data ($M : N$) in a training mini-batch on Market-1501.

### 4.4. Variant Evaluation

**Baseline evaluation.** To fully present the effectiveness of CamStyle, our baseline systems consist of 2, 3, 4, 5, 6 cameras for Market-1501 and 2, 3, 4, 5, 8 cameras for DukeMTMC-reID, respectively. In a system with 3 cameras, for example, the training and testing sets both have 3 cameras. In Fig. 7, as the number of cameras increases, the rank-1 accuracy increases. This is because 1) more training data are available and 2) it is easier to find a rank-1 true match when more ground truths are present in the database. In the full-camera (6 for Market-1501 and 8 for DukeMTMC-reID) baseline system, the rank-1 accuracy is 85.6% on Market-1501 and is 72.3% on DukeMTMC-reID.

**Vanilla CamStyle improves the accuracy of few-camera systems.** We first evaluate the effectiveness of the vanilla method (without LSR) in Fig. 7 and Table 1. We have two observations. First, in systems with 2 cameras, *vanilla CamStyle* yields significant improvement over the baseline CNN. On Market-1501 with 2 cameras, the improvement reaches +17.1% (from 43.2% to 60.3%). On DukeMTMC-reID with 2 cameras, the rank-1 accuracy is improved from 45.3% to 54.8%. This indicates that the few-camera systems, due to the lack of training data, are prone to over-fitting, so that our method exhibits an impressive system enhancement.

Second, as the number of camera increases in the system, the improvement of *vanilla CamStyle* becomes smaller. For example, in the 6-camera system on Market-1501, the improvement in rank-1 accuracy is only +0.7%. This indicates that 1) the over-fitting problems becomes less severe in this full system and that 2) the noise brought by CycleGAN begins to negatively affect the system accuracy.

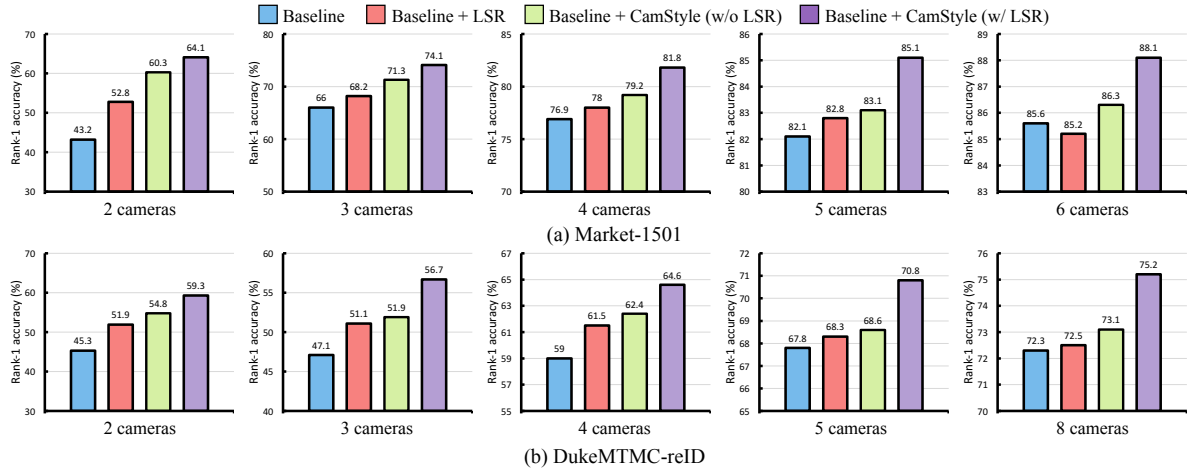

Figure 7. Comparison of different methods on Market-1501 and DukeMTMC-reID, *i.e.*, baseline, baseline+LSR, baseline+CamStyle *vanilla* (w/o LSR), baseline+CamStyle (w/ LSR). Rank-1 accuracy is shown. Five systems are shown, which have 2, 3, 4, 5, 6 cameras for Market-1501 and 2, 3, 4, 5, 8 cameras for DukeMTMC-reID, respectively. We show that CamStyle (with LSR) yields consistent improvement over the baseline.

| Training data | $\mathcal{L}_R$ | $\mathcal{L}_F$ | Rank-1 | mAP |
|---|---|---|---|---|
| Real | CrossE | None | 85.66 | 65.87 |
| Real | LSR | None | 85.21 | 65.60 |
| Real+Fake | CrossE | CrossE | 86.31 | 66.02 |
| Real+Fake | CrossE | LSR | **88.12** | **68.72** |

Table 1. Performance evaluation on Market-1501 using different loss functions. CrossE: Cross-entropy, LSR: Label smooth regularization [30].

| Method | Rank-1 | mAP |
|---|---|---|
| Baseline | 85.66 | 65.87 |
| Baseline+CamStyle (1+2) | 87.20 | 67.64 |
| Baseline+CamStyle (1+2+3) | 87.32 | 68.53 |
| Baseline+CamStyle (1+2+3+4) | 87.42 | 68.23 |
| Baseline+CamStyle (1+2+3+4+5) | 87.85 | 68.51 |
| Baseline+CamStyle (1+2+3+4+5+6) | **88.12** | **68.72** |

Table 2. Impact analysis of using different cameras for training CycleGANs on Market-1501. We adopt the 6-camera system. We start from using the 1st and 2nd cameras, and then gradually add other cameras for training CycleGANs.

| Method | RF+RC | RE | CamStyle | Rank-1 | mAP |
|---|---|---|---|---|---|
| Baseline |  |  |  | 84.15 | 64.10 |
|  | ✓ |  |  | 85.66 | 65.87 |
|  |  | ✓ |  | 86.83 | 68.50 |
|  |  |  | ✓ | 85.01 | 64.86 |
|  | ✓ | ✓ |  | 87.65 | 69.91 |
|  | ✓ |  | ✓ | 88.12 | 68.72 |
|  |  | ✓ | ✓ | 87.89 | 69.10 |
|  | ✓ | ✓ | ✓ | **89.49** | **71.55** |

Table 3. Comparison combinations between different data augmentation methods on Market-1501. **RF+RC**: random flip+random crop, **RE**: Random Erasing [49].

**LSR is effective for CamStyle.** As previously described, when tested in a system with more than 3 cameras, *vanilla CamStyle* achieves less improvement than the 2-camera system. We show in Fig. 7 and Table 1 that using the LSR loss on the fake images achieves higher performance than cross-entropy. As shown in Table 1, using cross-entropy on style-transferred data improves the rank-1 accuracy to 86.31% under full-camera system on Market-1501. Replacing cross-entropy with LSR on the fake data increases the rank-1 accuracy to 88.12%.

In particular, Fig. 7 and Table 1 show that using LSR alone on the real data does not help much, or even decrease the performance on full-camera systems. *Therefore, the fact that CamStyle with LSR improves over the baseline is not attributed to LSR alone, but to the interaction between LSR and the fake images.* By this experiment, we justify the necessity of using LSR on the fake images.

**The impact of using different cameras for training camera-aware style transfer models.** In Table 2, we show that as using more cameras to train camera-aware style transfer models, the rank-1 accuracy is improved from 85.66% to 88.12%. Particularly, our method obtains +1.54% improvement in rank-1 accuracy even only using the 1th and 2th camera to train camera-aware style transfer model. In addition, when training cameras style transfer models with using 5 cameras, it has the rank-1 accuracy of 87.85%, which is 0.27% lower than of using 6 cameras. This shows that even using a part of the cameras to learn camera-aware style transfer models, our method can yield approximately equivalent results to using all the cameras.

**CamStyle is complementary to different data augmentation methods.** To further validate the CamStyle, we comparison it with two data augmentation methods, random flip + random crop (RF+RC) and Random Erasing (RE) [49]. RF+RC is a common technique in CNN training [17] to improve the robustness to image flipping and object translation. RE is designed to enable invariance to occlusions.

| Method | Rank-1 | mAP |
|---|---|---|
| BOW [42] | 34.40 | 14.09 |
| LOMO+XQDA [20] | 43.79 | 22.22 |
| DNS [39] | 61.02 | 35.68 |
| IDE [43] | 72.54 | 46.00 |
| Re-rank [48] | 77.11 | 63.63 |
| DLCE [45] | 79.5 | 59.9 |
| MSCAN [18] | 80.31 | 57.53 |
| DF [40] | 81.0 | 63.4 |
| SSM [1] | 82.21 | 68.80 |
| SVDNet [29] | 82.3 | 62.1 |
| GAN [47] | 83.97 | 66.07 |
| PDF [28] | 84.14 | 63.41 |
| TriNet [11] | 84.92 | 69.14 |
| DJL [19] | 85.1 | 65.5 |
| IDE* | 85.66 | 65.87 |
| **IDE*+CamStyle** | 88.12 | 68.72 |
| **IDE*+CamStyle+RE** [49] | **89.49** | **71.55** |

Table 4. Comparison with state of the art on the Market-1501 dataset. IDE* refers to improved IDE with the training schedule in this paper. **RE**: Random Erasing [49].

| Method | Rank-1 | mAP |
|---|---|---|
| BOW+kissme [42] | 25.13 | 12.17 |
| LOMO+XQDA [20] | 30.75 | 17.04 |
| IDE [43] | 65.22 | 44.99 |
| GAN [47] | 67.68 | 47.13 |
| OIM [37] | 68.1 | 47.4 |
| APR [21] | 70.69 | 51.88 |
| PAN [46] | 71.59 | 51.51 |
| TriNet [11] | 72.44 | 53.50 |
| SVDNet [29] | 76.7 | 56.8 |
| IDE* | 72.31 | 51.83 |
| **IDE*+CamStyle** | 75.27 | 53.48 |
| **IDE*+CamStyle+RE** [49] | **78.32** | **57.61** |

Table 5. Comparison with state of the art on DukeMTMC-reID. IDE* refers to improved IDE with the training schedule described in this paper. **RE**: Random Erasing [49].

As show in Table 3, rank-1 accuracy is 84.15% when no data augmentation is used. When only applying RF+RC, RE, or CamStyle, rank-1 accuracy is increased to 85.66%, 86.83% and 85.01%, respectively. Moreover, if we combine CamStyle with either RF+RC or RE, we observe consistent improvement over their separate usage. The best performance is achieved when the three data augmentation methods are used together. Therefore, while the three distinct data augmentation techniques focus on different aspects of CNN invariance, our results show that, CamStyle is well complementary to the other two. Particularly, combining these three methods, we achieve 89.49% rank-1 accuracy.

### 4.5. Comparison with State-of-the-art Methods

We compare our method with the state-of-the-art methods on Market-1501 and DukeMTMC-reID in Table 4 and Table 5, respectively. First, using our baseline training strategy, we obtain a strong baseline (IDE*) on both datasets. Specifically, IDE* achieves 85.66% for Market-1501 and 72.31% for DukeMTMC-reID in rank-1 accuracy. Compared with published IDE implementations [29, 47, 43], IDE* has the best rank-1 accuracy on Market-1501.

Then, when applying CamStyle on IDE*, we obtain competitive results compared with the state of the art. Specifically, we achieve **rank-1 accuracy = 88.12% for Market-1501, and rank-1 accuracy = 75.27% for DukeMTMC-reID.** On Market-1501, our method has higher rank-1 accuracy than PDF [28], TriNet [11] and DJL [19]. On the other hand, the mAP of our method is slightly lower than TriNet [11] by 0.42% on Market-1501 and lower than SVDNet [29] by 3.32% on DukeMTMC-reID.

Further combining CamStyle with Random Erasing data augmentation [49] (RF+RC is already implemented in the baseline), our final rank-1 performance arrives at **89.49%** for Market-1501 and **78.32%** for DukeMTMC-reID.

## 5. Conclusion

In this paper, we propose CamStyle, a camera style adaptation method for deep person re-identification. The camera-aware style transfer models are learned for each pair of cameras with CycleGAN, which are used to generate new training images from the original ones. The real images and the style-transferred images form the new training set. Moreover, to alleviate the increased level of noise induced by CycleGAN, label smooth regularization (LSR) is applied on the generated samples. Experiments on the Market-1501 and DukeMTMC-reID datasets show that our method can effectively reduce the impact of over-fitting, and, when combined with LSR, yields consistent improvement over the baselines. In addition, we also show that our method is complementary to other data augmentation techniques. In the feature, we will extend CamStyle to one view learning and domain adaptation.

## 6. Acknowledgements

This work is supported by the National Nature Science Foundation of China (No. 61572409, No. U1705286 & No. 61571188), Fujian Province 2011Collaborative Innovation Center of TCM Health Management, Collaborative Innovation Center of Chinese Oolong Tea Industry-Collaborative Innovation Center (2011) of Fujian Province, Fund for Integration of Cloud Computing and Big Data, Innovation of Science and Education, the Data to Decisions CRC (D2D CRC) and the Cooperative Research Centres Programme. Yi Yang is the recipient of a Google Faculty Research Award. Liang Zheng is the recipient of a SIEF STEM+ Bussiness fellowship. We thank Wenjing Li for encouragement.